\documentclass{AISB2008}
\pdfoutput=1
\usepackage{times}
\usepackage{graphicx}
\usepackage{latexsym}
\usepackage{url}

\begin{document}
	
\title{Exploring Novel Game Spaces with Fluidic Games}
	
\author{Swen E. Gaudl, Mark J. Nelson, Simon Colton, Rob Saunders, \\
	Edward J. Powley, Peter Ivey, Blanca P\'erez Ferrer, Michael Cook \institute{MetaMakers
    Institute, Falmouth University, Cornwall, UK, website:
    http://metamakersinstitute.com} }
	
\maketitle
\bibliographystyle{AISB2008}
	
\begin{abstract}
With the growing integration of smartphones into our daily lives, and their
increased ease of use, mobile games have become highly popular across all
demographics. People listen to music, play games or read the news while in
transit or bridging gap times. While mobile gaming is gaining popularity, mobile
expression of creativity is still in its early stages. We present here a new
type of mobile app -- fluidic games -- and illustrate our iterative approach to
their design. This new type of app seamlessly integrates exploration of the
design space into the actual user experience of playing the game, and aims to
enrich the user experience. To better illustrate the game domain and our
approach, we discuss one specific fluidic game, which is available as a
commercial product. We also briefly discuss open challenges such as player
support and how generative techniques can aid the exploration of the game space
further.
\end{abstract}
	
\section{INTRODUCTION}

Mobile games have become a large sub-market of the global games industry to the
extent that many companies specialise in developing mobile apps. This large
market share is due to the pervasive nature of smartphones and the low technical
hurdle of obtaining mobile apps from app stores. Mobile games are ever-present
and are consumed by nearly all demographic groups, including many not reached by
traditional desktop or console games. However, designing games is still
dominated by desktop applications and commercial production tools such as Unity
or Unreal, which have a steep learning curve and require software development
skills. While there have been introductory tools developed to teach foundations
of programming and game design to novices and to support STEM education,
including programs such as Kodu or Scratch, these are still desktop-centric and
require serious time investment to develop games. Separate from game creation
tools, there is also a category of apps dubbed casual creators
\cite{compton2015}, which allow users to design digital toys. Casual creators
differ from game design tools as they are more centred around creative personal
expression than the design of a consistent game.
	
We present here a new type of mobile application which positions itself between
a mobile game, a game design tool and a casual creator. We call these apps
\emph{fluidic games} \cite{mici}. The rest of the paper is organised as
follows. We first introduce the notion of fluidic games and how they differ from
current games and game design tools. We will then more closely examine one
fluidic game that we have produced, and discuss the process of developing
it. Having introduced and elaborated our game, we will present open research
questions and future directions.
	
\section{FLUIDIC GAMES}

Fluidic games, in contrast to games as normally conceived, contain a subspace of
different games that can be designed within the app itself. Thus, a fluidic game
is not just one game, which is a single point in game-design space, but an
entire design space of games through which the player can move and explore.  This
concept of expanding a single game into a game design space offers an increase
in replayability beyond the state of art in commercial games and is intended to
also foster more creative expression of the player, who both plays and creates
games. Currently, focusing on apps used directly on hand-held devices, casual
creation for games is limited to skinning games, designing levels in an existing
game world, and/or programming through products like Scratch Jnr. On other
platforms, players came ``mod'' software to modify the game logic
\cite{gaudl2013behaviour} which is both limiting and complicated.
	
The idea we present here was developed through continuous designer and user
feedback and driven by research in computational creativity \cite{colton:ecai12}
to expand a single game into a game subspace which contains a coherent set of
attributes uniting games in that subspace. As a starting point, we built a
larger game-design space, to be navigated via software called \emph{Gamika
  Technologies} \cite{mici,powley2016automated}, and looked at restricting it to
more cohesive subspaces, which share common dimensions of the Gamika space and
thus reduce the individual on-device design spaces by focusing on specialised
interfaces and automating generative aspects to navigate the desired subspace.
	
Despite all games being 2D and physics-based, the Gamika space is heterogeneous,
with very different kinds of games available within its parameters; some
puzzle-like, others meditative, others arcade-style action, etc. Within this
heterogenous design space, there are cohesive subspaces. Games within such a
cohesive subspace have a larger overlap in common features which renders the
navigating within the design subspace closer to transitioning between game
variants, with more understandable relationships between the impact of their
parameters on gameplay behaviour (though often still with unexpected and emergent
aspects).  Once such a design subspace is identified, the research question
shifts towards understanding the variation in games is affords well enough to
build user-interface and generative components that match with its salient
features, and employing those to build an enjoyable, mixed-initiative app for
designing (and playing) games or levels in that subspace. We describe below a
subspace and the corresponding mobile game-design app, Wevva.
	
\subsection{Development of a Fluidic Game}

As an anchor point for our fluidic game, we started with a single point in the
design subspace and picked one single concept to create a stand-alone
product. This product is the Let It Snow app, our first commercial game
developed using Cillr, our in-house mobile design app to navigate the full
Gamika space, and the centrepoint around which our new fluidic game Wevva
unfolds. After arriving at a finished game, we expanded the space around Let It
Snow similar to a sculptor working with clay, by adding and shaping material in
an iterative process. By doing so, we opened up more and more of the Gamika
space around that single point. This process was driven by user testing and
design sessions, focusing and expanding the space towards areas of interest to
the designers and users. As part of this process, we carried our multiple game
jams and design sessions which we detail below.
	
One key aspect of fluidic games is to focus on the possibility of extremely
short game design sessions and what information the user needs to navigate the
game space effectively. With the Wevva, app, for instance, it is possible to
design a new game in ten minutes or less. Thus, navigating the cohesive subspace
and arriving at an interesting game in a short amount of time is key to a
fluidic game.
	
\subsection{Let It Snow}

Let It Snow is the first game from the MetaMakers Institute and available for
iOS. The game, which is a regular game (not a fluidic game), was developed using
Cillr, our in-house tool to navigate the Gamika design space. It was entirely
designed on a mobile device; after the design phase, it was exported from Cillr
and polished to be publicly released in Apple's App Store in under two weeks of
development. Although superficially a casual game requiring simple tap and swipe
interactions, it is designed as an easy to play but hard to master game
requiring the player to discover and employ different strategies.

\begin{figure}
	\centering
        \includegraphics[height=0.25\textheight]{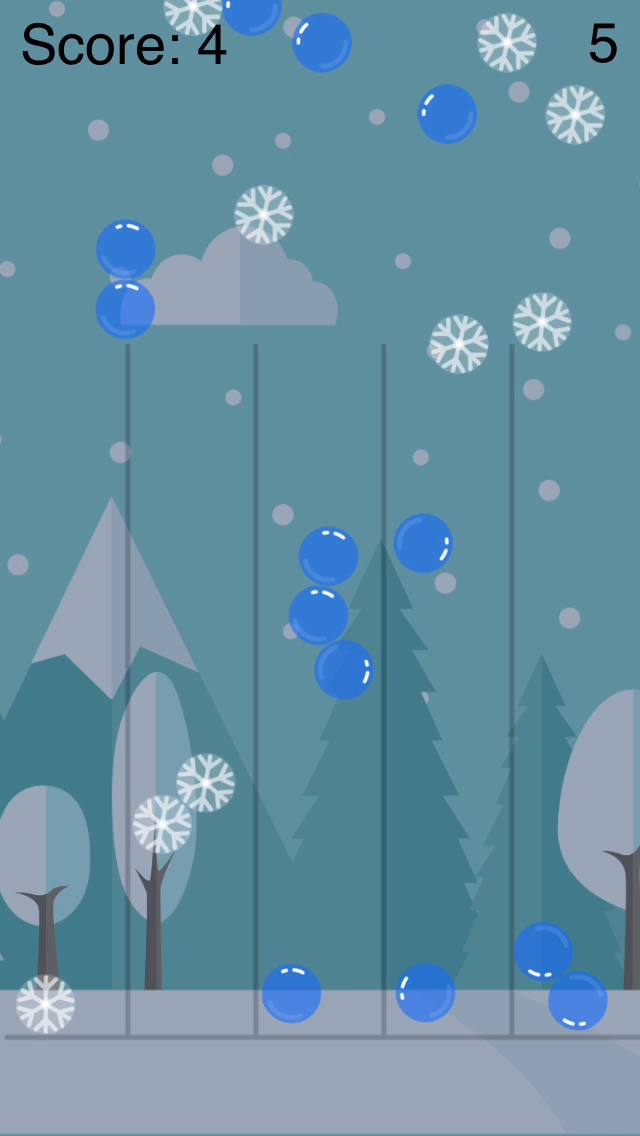}
	\caption{\emph{Let It Snow} game interface showing the score of the player in the top left and the elapsed time in the top right.}
	\label{fig:lis-interface}
\end{figure}

The rules are very simple, but getting good at it takes some
effort. Additionally, if the players want to do well in a particular game, they
require both high concentration for the duration of the game and some degree of
luck, due to random spawning of game elements and emergent properties of the
physics simulation making each game different. In Figure~\ref{fig:lis-interface} you
can see the main screen of the game.

The game rules are as follows. Snow and rain pour down from the top of the
screen (as white and blue balls respectively). When four or more white balls
cluster together, they burst, and the player gains a point for each in the
cluster. Each white ball that explodes is replaced by a new one spawned at the
top, with a maximum of 20 on screen at any one time. Likewise with blue balls,
except the player loses a point for each that explodes. Players can interact
with the game by tapping blue balls to explode them, losing one point in doing
so.

While the game rules are straightforward, we have found it to be difficult and
require puzzle-solving strategies as well as quick reactions. There is a grid
structure which collates the balls into bins, and the best way to play the game
involves trapping the blue balls in groups of twos and threes at the bottom,
while the whites are exposed and are continually refreshed through cluster
explosions. Occasionally, when all blues are trapped in small clusters, only
whites will spawn, which is akin to snowing (hence the game’s name) and is a
particularly pleasing moment to aim for.

After having released Let It Snow, we expanded the design space to offer players
a way to explore parts of the Gamika desgn space around Let It Snow. To do so,
we exposed parameters to user modification, opening up a controlled subspace
around Let It Snow without venturing too far from this anchor point. Focusing in
this way offers the player the possibility to not only create entirely new
games, but also to alter and modify Let It Snow, making it harder or easier as
they see fit. This controlled exploration of the space around Let It Snow should
lower the cognitive load compared to allowing free-form design within the full
Gamika design space, as the games within the subspace all share a large set of
commonalities.

\subsection{Wevva}
Again using Cillr to navigate the full Gamika space, we produced three
variations of Let It Snow called Rain Rain, Jack Frost and Slush Slosh, each
requiring different tactics and skills.

\begin{figure}
	\centering
	\includegraphics[height=0.38\textheight]{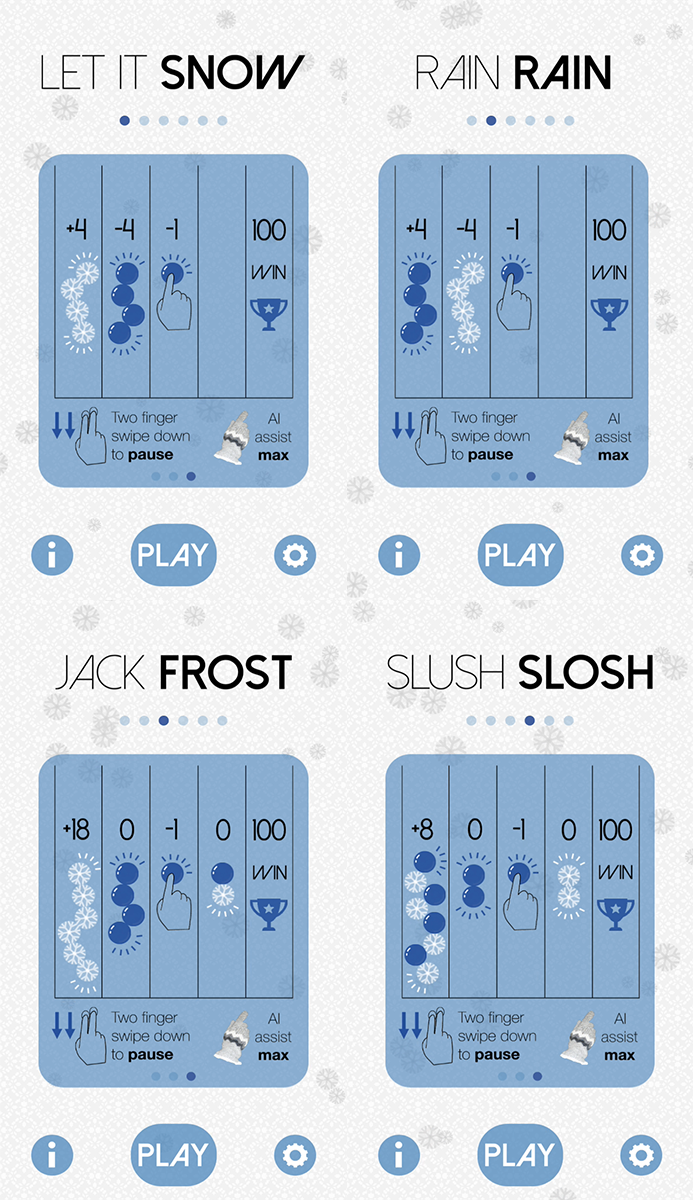}
	\caption{The \emph{Wevva} interface showing the different rules of all four included games which form the foundation of the game space.}
	\label{fig:snowfull-instructions}
\end{figure}

Those four games populated the initial subspace forming the base of our fluidic
game. They will be released as part of our iOS game entitled \emph{Wevva}
(Figure~\ref{fig:snowfull-instructions}). This app further includes two aspects
that are not common in casual games: (a) an AI player for each game that can
assist novice players, and (b) a design screen enabling players to navigate to
game variants within the fluidic subspace in a semi-random way, and tweak them
to get balanced variations. This is more extensive than a level-design screen
(which is somewhat more common in existing games), as not only the physical
layout of levels can be changed, but the various game rules and mechanics,
including aspects such as physics and scoring.

\begin{figure}
	\centering
	\includegraphics[height=0.38\textheight]{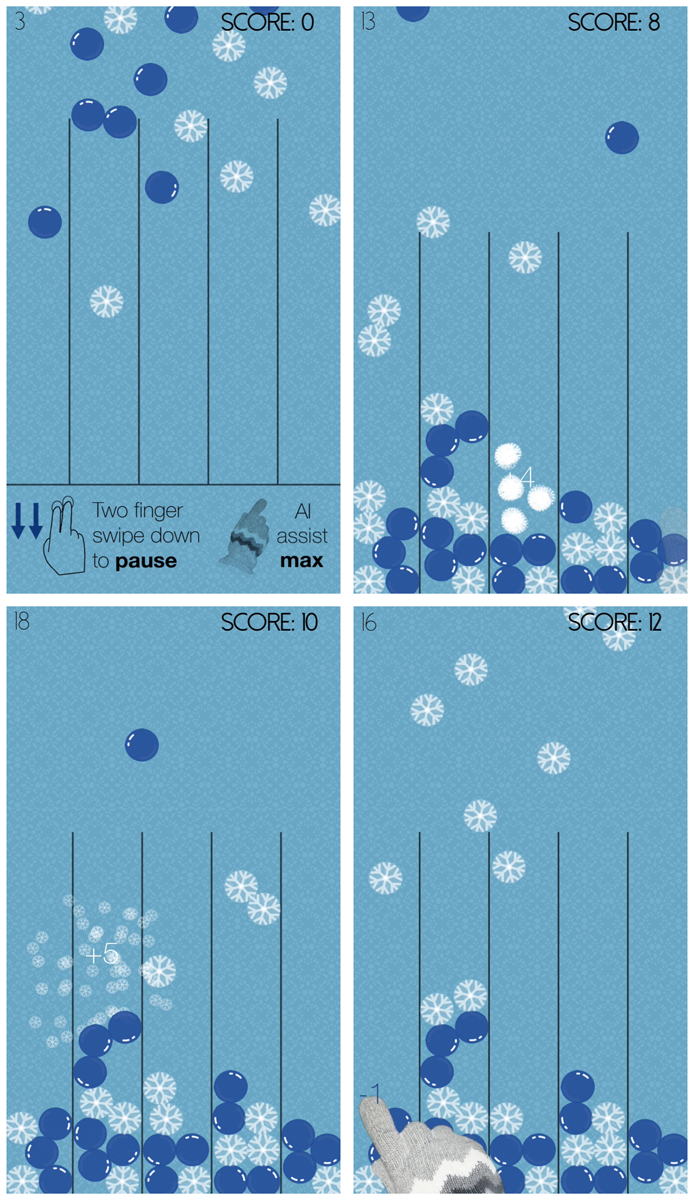}
	\caption{The \emph{Wevva} interface showing the different rules of all four included games which form the foundation of the game space.}
	\label{fig:snowfull-gameplay}
\end{figure}

The AI player is tuned differently for all four games and appears on-screen as a
gloved hand to support the player by tapping balls which would reduce the score
in the individual game (Figure~\ref{fig:snowfull-gameplay}, bottom right). With
full support of the AI player, players can concentrate on higher level
strategies, resulting in a reduced difficulty of the game as the player do not
need to switch their focus. To offer a more rewarding and demanding gameplay the
app provides a slider to change the level of AI assistance.  At 50\%, the
support should feel like having an in-game partner helping out. At 0\%, the game
plays quite different, as the AI player is not helping any more. At this
setting, the player has to continuously switch between stopping clusters of
unwanted balls from forming and also pursuing a high-level strategy and
connecting scoring clusters; this is also the hardest way to play the four
included games without making other adjustments such as increasing the speed.

\begin{figure}[t]
	\centering
        \includegraphics[height=0.38\textheight]{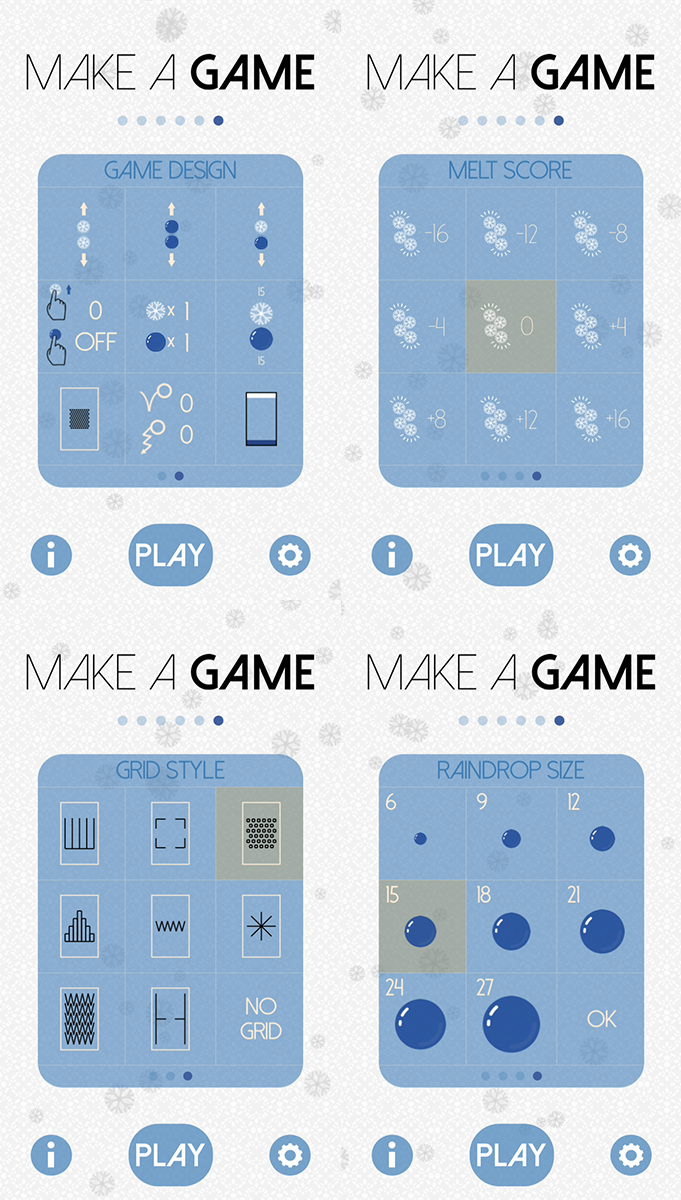}
	\caption{\emph{Wevva} design interface}
	\label{fig:snowfull-design}
\end{figure}

The design screen (Figure~\ref{fig:snowfull-design} top left) exposes the
following nine facets of the game design space to the player: (a) the cluster sizes at
which balls explode, (b) the scores attached to clusters exploding
and the player tapping, (c) the ball sizes, (d) the allowed maximum number of balls
of each type, (e) the grid design, (f) physical properties of the
environment, namely bounciness and noise, (g) spawning regions for both ball types, (h) scoring regions for balls exiting the screen, and (i) what happens
when the player taps the balls -- both actions and scoring consequences.  In
Figure~\ref{fig:snowfull-design} in the top right, the score panel for the
previous facet (a) is shown and in the bottom left the grid choices, to
illustrate the depth of the design space. There is a random generation button
which will set these parameters in a varied way, but designed so that the
clustering explosion are balanced in terms of their expected impact on the
score. We achieved this by running online simulations of novice players and
recording the number of times that clusters of each size and type
occurred. Initial experiments with the design screen have indicated that the
space exposed by the above parameters, while vast, does not contain hugely
varied game types, hence the space seems manageable.  However, we have used it
to make games which differ substantially from the four preset games, e.g.,
involving juggling balls, or trapping and tapping them, etc.

\section{GAME JAMS, DESIGN SESSIONS AND \\ OPEN CHALLENGES}

While expanding upon the initial Let It Snow app to a coherent fluidic game, we
conducted a number of internal brainstorming and design sessions. During those
sessions, we discussed which aspects of the subspace would be of more interest
to users. During this process, we also focused on the ability to create games
quickly. With a coherent design subspace, we could limit the number of
parameters exposed to players which reduce the design time of new games
significantly. Because Wevva focuses on users who are not necessarily in an
office or at home, it must be possible to quickly design games.
	
Having designed the initial version of the app through in-house playtesting, we
conducted a first external user test with 65 members of Girlguiding
Cornwall's Brownie programme (girls aged 5-9), who visited Falmouth
University as part of a larger Girls Can Code event on 18th February 2017. We conducted two sessions, with 35 users in the first and 30 in the second,
with groups of 2 to 3 children sharing an iPad. The first part was consisted of
a conventional playtest, where we demonstrated \emph{Let It Snow} and the other
three included games to give them a starting point. As a next step, we asked
them to play those games for about ten minutes. Subsequently, we introduced them
to the design interface and gave them about an hour to explore the design space
and design their own games and share them with other participants. It is
worth noting that the app was not designed with young children as the
target audience. However, we found that the girls had no problem using the app and they navigated the game space without having to read or understand complex
instructions.

The playtest of the four built-in games produced largely negative results. These
games are puzzle-oriented, requiring the player to be patient and come up with a
winning strategy; but here, very few were able to discover a winning strategy.
On the other hand, the design experience was more successful. Observations
during those two sessions showed that the participants had no problems
navigating and using the app and were able to navigate the fluidic design
game. We are currently evaluating the gathered data, but a first result is that
all participants were able to design their own games using the app. Most of
those games were focused on simpler mechanics such as rapid tapping or using the
controller to remove balls from the screen quickly. Still, bearing in mind that
the participants were very young, we believe that these initial results
demonstrate how easy to use the app is, and the low barriers to entry.
	
We conducted two further game jams with 40 members of Girlguiding Cornwall's
Guides programme (i.e., girls aged 10-14), who visited Falmouth University on
23rd February 2017 as part of a larger Girls Can Code event. We repeated the
same structure as in the first game jams, starting with an introduction to the
four included games and a ten-minute game playing session to familiarise them
with the game, its controls and mechanics. In contrast to the younger
participants from the first two sessions, the participants in these sessions
spent more time playing the four included games. They also approached them more
closely to our expectations, probing and trying out different strategies. They
were then given an introduction to the design space, and, as with the first
group, given the possibility to explore the design space by creating and sharing
their own games. Initial observations showed that the designed games sometimes
still focused on fast tapping game mechanics, but we also saw a wider range of
games which required more sophisticated strategies. Thus, more advanced games
were developed. The older participants also focused on games which could not be
beaten instantly and ventured further into the fluidic design space.
	
Following this set of four user test sessions, we adjusted menu structures and
support texts in response to observations of where players had difficulty, and
comments they made on a feedback form. We also found bugs within the app which
we were unaware of. Due to the complex nature of the design space and the way a
designer can approach the same point in the game space from different angles, it
is quite difficult to conduct bullet proof tests. Thus, user testing sessions
not only give meaningful feedback about the game, but also help identity less
obvious issues.
	
\subsection{Open Challenges}

When the initial app Let It Snow was released, it came without an AI support
system embedded to aid the user. The game does look like a typical casual game
but requires a lot of focus and determination to achieve mastery. The
requirement was by design but makes it hard for new users to understand the game
mechanics or keep motivated to play the game. With the new fluidic game Wevva,
we shipped a different AI support system for each of the four original
games. The AI can be adjusted by scaling the support from not helping the user
at all to near perfect help. The AI was included for two purposes. Firstly, to
reduce the number of quick reflex-like actions the players have to perform, so
they are put in charge of controlling the difficulty. Secondly, to help the
generator play and test new games while it is exploring the space to see if
those games are playable.
	
During our game jams with the younger participants, we observed that, since they
did not fully understand the hidden game mechanics, the AI system was
misunderstood. The participants believed that the AI was not aiding but playing
against the user, since its immediate effect was to explode bubbles that caused
a loss of a point. This led to some initial frustration. With the older
participants, the AI assistant's behaviour was better understood. Overall, we
found that the AI support needs to introduce its intention in some way to the
user to be correctly conceptualised.
	
The second purpose of the AI, to aid the generator, proved to be more difficult
and is still an open challenge. The game space of our fluidic game is coherent
and less open than the full Gamika space but still requires the AI to employ
different techniques and sometimes employ unconventional strategies. As the AI
has to run within the requirements of the mobile device, classical learning
approaches such as neural networks or large scale simulation are harder to
employ. We are currently investigation more general AI player approaches or
offering an interface so that players can develop automated players.
	
\section{CONCLUSIONS}

We presented our approach for exploring coherent game spaces to achieve a new
type of mobile application, namely fluidic games, in which players can not only
play included games, but modify them and design wholly new games nearby in the
design space. We describe our approach and the process of carving out the
suitable subspace around a specific initial game, Let It Snow, to reach the
first fluidic game, Wevva. We also describe our design approach, integrating
participants through game james, and discussed open challenges for fluidic games
such as player support, and AI players for traversing the game space. We believe
that fluidic games offer enormous potential to open up game design to large
numbers of people, and also to highlight many interesting research challenges.
	
\ack This work is funded by EC FP7 grant 621403 (ERA Chair: Games Research
Opportunities). We are very grateful for the feedback provided by our alpha/beta
testers.
	
\bibliography{main,aisb}
	
\end{document}